\documentclass{article}
\usepackage{ijcai21}

\usepackage{times}

\usepackage{soul}
\usepackage{url}
\usepackage[hidelinks]{hyperref}
\usepackage[utf8]{inputenc}
\usepackage[small]{caption}
\usepackage{graphicx}
\usepackage{amsmath}
\usepackage{booktabs}
\usepackage{amsmath}
\usepackage{amsthm}
\usepackage{booktabs}
\usepackage{algorithm}
\usepackage{algorithmic}
\usepackage{amsfonts}
\usepackage{makecell}
\usepackage{multirow}
\usepackage{color}
\usepackage{float}
\title{Low Resolution Information Also Matters: Learning Multi-Resolution Representations for Person Re-Identification}

\author{
Guoqing Zhang$^1$\and
Yuhao Chen$^2$\and
Weisi Lin$^{1}$\footnote{Corresponding Author}\and
Arun Chandran$^3$\and
Xuan Jing$^3$\\
\affiliations
$^1$Nanyang Technological University, Singapore\\
$^2$Nanjing University of Information Science and Technology, China\\
$^3$Singapore Telecommunications Limited, Singapore\\
\emails
guoqing.zhang@ntu.edu.sg,
chinayhchen@gmail.com,
wslin@ntu.edu.sg,\\
\{arunkumar.chandran, xuan.jing\}@ncs.com.sg
}

\begin{document}

\maketitle

\begin{abstract}
As a prevailing task in video surveillance and forensics field, person re-identification (re-ID) aims to match person images captured from non-overlapped cameras. In unconstrained scenarios, person images often suffer from the resolution mismatch problem, i.e., \emph{Cross-Resolution Person Re-ID}. To overcome this problem, most existing methods restore low resolution (LR) images to high resolution (HR) by super-resolution (SR). However, they only focus on the HR feature extraction and ignore the valid information from original LR images. In this work, we explore the influence of resolutions on feature extraction and develop a novel method for cross-resolution person re-ID called \emph{\textbf{M}ulti-Resolution \textbf{R}epresentations \textbf{J}oint \textbf{L}earning} (\textbf{MRJL}). Our method consists of a Resolution Reconstruction Network (RRN) and a Dual Feature Fusion Network (DFFN). The RRN uses an input image to construct a HR version and a LR version with an encoder and two decoders, while the DFFN adopts a dual-branch structure to generate person representations from multi-resolution images. Comprehensive experiments on five benchmarks verify the superiority of the proposed MRJL over the relevent state-of-the-art methods.
\end{abstract}

\section{Introduction}

Person re-identification (re-ID) is a retrieval task of recognizing the same person across images from non-overlapped cameras, which has attracted increasing attention in computer vision community due to its wide application prospects in video surveillance and forensics field \cite{1}. Nevertheless, person re-ID remains a challenge due to some complicated visual variations in real scenarios such as viewpoint, illumination, person pose and background clutter.

Most existing re-ID methods focus on the designment of feature extraction networks or matching distance metrics on the basis of the assumption that all captured images share similar and sufficiently high resolutions. However, this assumption only exists in an absolutely ideal condition. In real and unconstrained scenarios, affected by some objective factors such as shooting distance and camera pixels, the captured images have variable resolutions. The problem of matching person images with variable resolutions is defined as \emph{Cross-Resolution Person Re-ID}.

\begin{figure}[t]
\centering
\includegraphics[width=0.9\columnwidth]{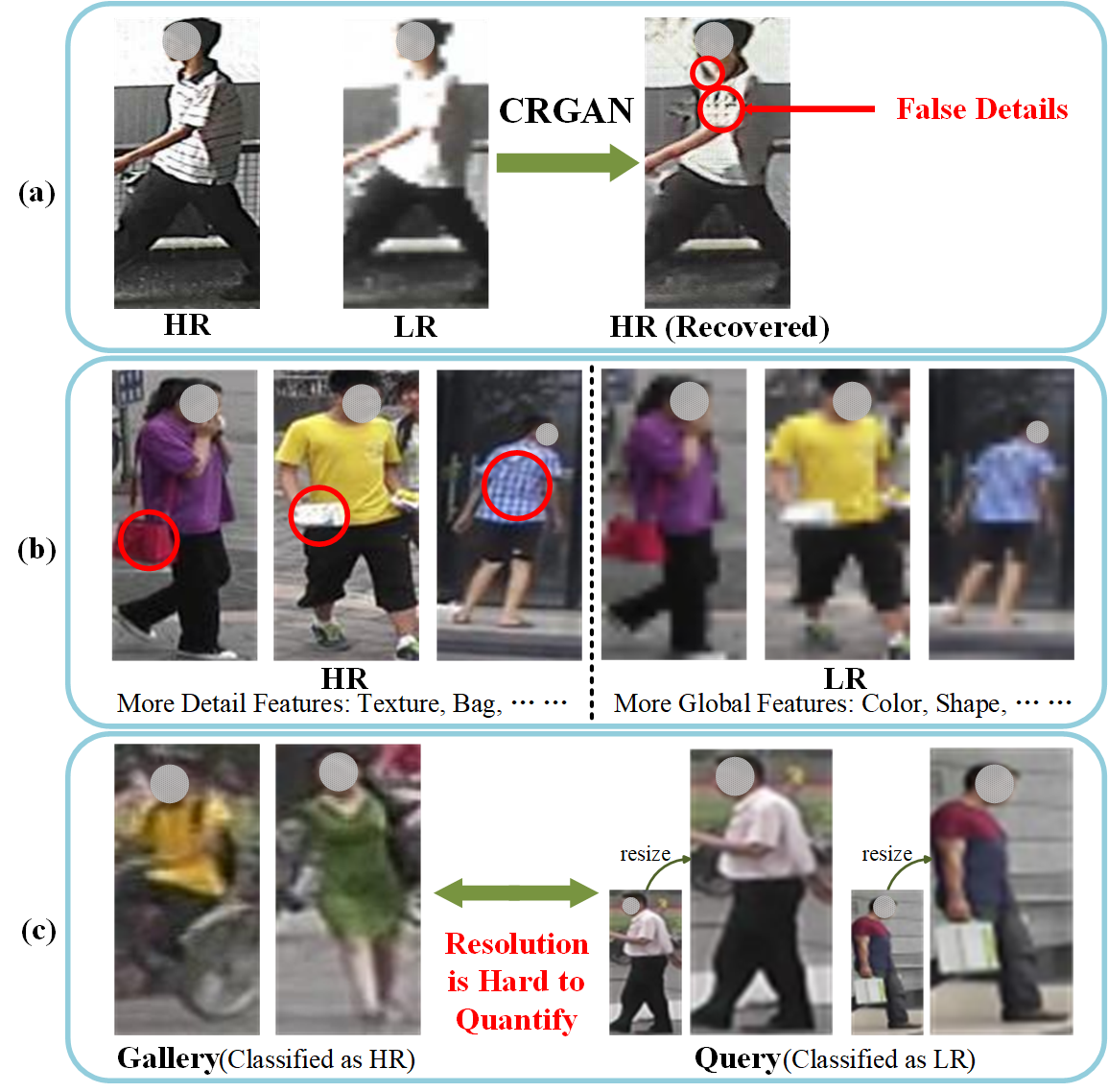}
\caption{Several shortages and limitations of existing cross-resolution person re-ID methods. (a) The recovered HR image with CRGAN may contain a few false details which may mislead the feature extraction. (b) HR images contain plenty of detail information, such as texture and bags, while current methods have not yet exploited complementary global features of LR images. (c) In some cases, resolution is hard to quantify with image pixel size. Some gallery images classified as HR even have worse visual quality than query images classified as LR.}
\label{fig1}
\end{figure}

Recently, a few researchers have paid attention to this problem and proposed some high-performance methods which can be mainly divided into two categories: 1) traditional methods utilizing metric learning or dictionary learning \cite{8,31} and 2) deep learning methods applying super-resolution (SR) technology to restore LR images to HR images, which are most commonly used in cross-resolution person re-ID \cite{10,11,12}. 
However, there are still problems in existing SR based methods, just as illustrated in Figure \ref{fig1}. 1) Existing methods mainly focus on recovering higher resolution images and extract HR feature representations. Although the complementary details generated by SR give person images better visual quality, these details may not be real in person appearances. Therefore, in some cases, the features extracted from these generated HR images are not discriminative enough to match correct persons. 2) Although local details are lost in LR images, LR images still can provide some global information, such as body shape and color, as evidenced in the studies of pyramid representation of images \cite{44}. These LR features can complement HR features which may be false details, but all existing methods neglect this useful information. 3) Most existing methods process gallery and query images with different strategies separately, because they tacitly approve all gallery images as HR and all query images as LR. However, in some practical scenarios, the resolutions of gallery or query images are not clearly divided, making it difficult to quantify the image as HR or LR.

In this paper, we investigate the influence of resolution on feature extraction and find that a neural network focuses on more local details in HR person images but more global features in LR person images. Inspired by this, we propose a novel \emph{\textbf{M}ulti-Resolution \textbf{R}epresentations \textbf{J}oint \textbf{L}earning} (\textbf{MRJL}) for cross-resolution person re-ID, which fully utilizes the detail information in HR and complementary information in LR. Our MRJL is made up of two sub-networks named as Resolution Reconstruction Network (RRN) and Dual Feature Fusion Network (DFFN). The RRN adopts a multi-kernel encoder to encode the input image into a feature map, and then applies two different decoders to restore the feature map to HR image and LR image, respectively. The DFFN utilizes a dual-branch structure based on the PCB method \cite{2} to generate person representations from multi-resolution images. It is worth noting that in the testing phase, our MRJL does not need to know the resolution of input image and treats images with different resolutions equally.

The contributions of our work are summarized as follows: 1) As far as we know, it is the first work to detailly explore the influence of resolution on feature extraction in person re-identification. 2) A novel method named as Multi-Resolution Representations Joint Learning (MRJL) is proposed for cross-resolution person re-ID, which fully utilizes features contained in different resolutions.

\section{Related Work}
\subsection{Person Re-ID}
In the past decade, a variety of high-performance methods have sprung up in the field of person re-ID. Most of these existing methods attempt to extract more discriminative features and overcome the difficulties such as pose changes and background clutter. For instance, some methods \cite{2,3} divide person image into several parts and extract local features which contain more discriminative details. Nevertheless, pose changes will affect the feature alignment. To address this problem, some excellent methods adopt pose-transferable GAN \cite{23} or pose estimation \cite{15} to enhance the robustness of network towards pose variations. To attenuate background clutter, some methods apply semantic parsing \cite{4} to remove backgrounds or apply attention mechanism \cite{22} to train the network to focus on more informative areas. However, all above methods are limited in practical use due to the incapability of adaptation to variable image resolutions in unconstrained scenarios.

\begin{figure*}[t]
\centering
\includegraphics[width=0.8\textwidth]{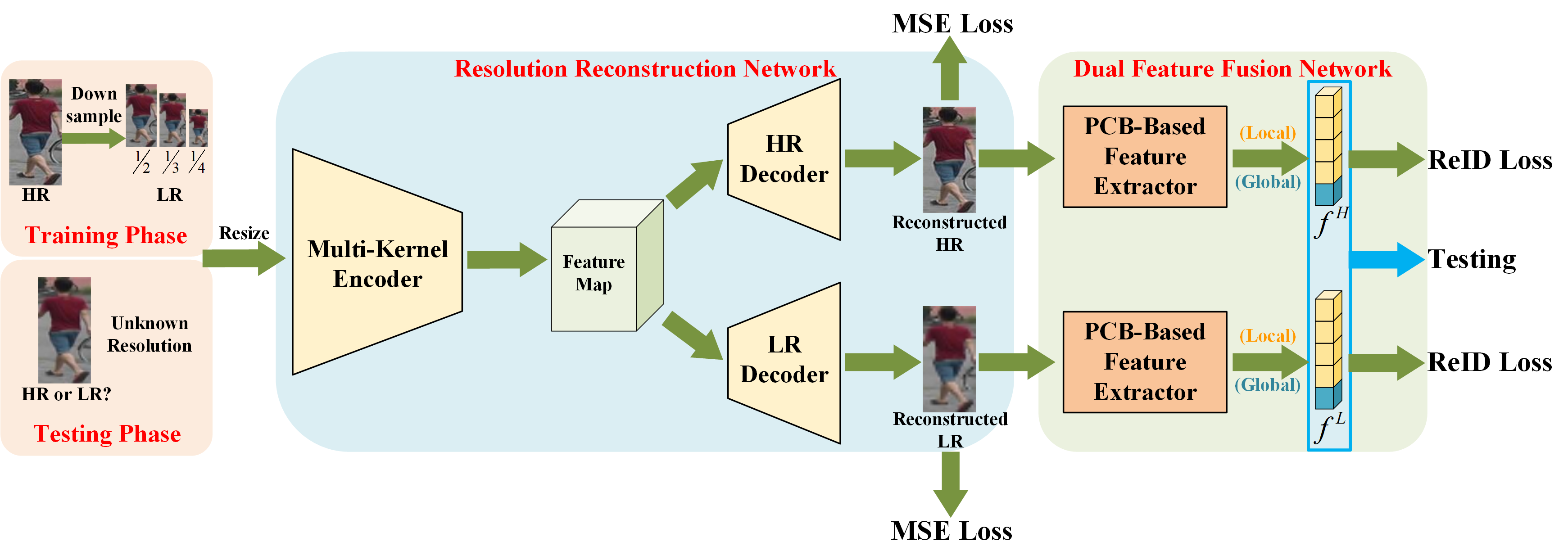}
\caption{The architecture of the proposed MRJL. This framework consists of two jointly trained sub-networks, Resolution Reconstruction Network (RRN) and Dual Feature Fusion Network (DFFN). The former is tasked to reconstruct input images into two versions with different resolutions, while the latter is used to extract feature representations from the generated HR and LR images.}
\label{fig2}
\end{figure*}

\begin{figure}[t]
\centering
\includegraphics[width=0.9\columnwidth]{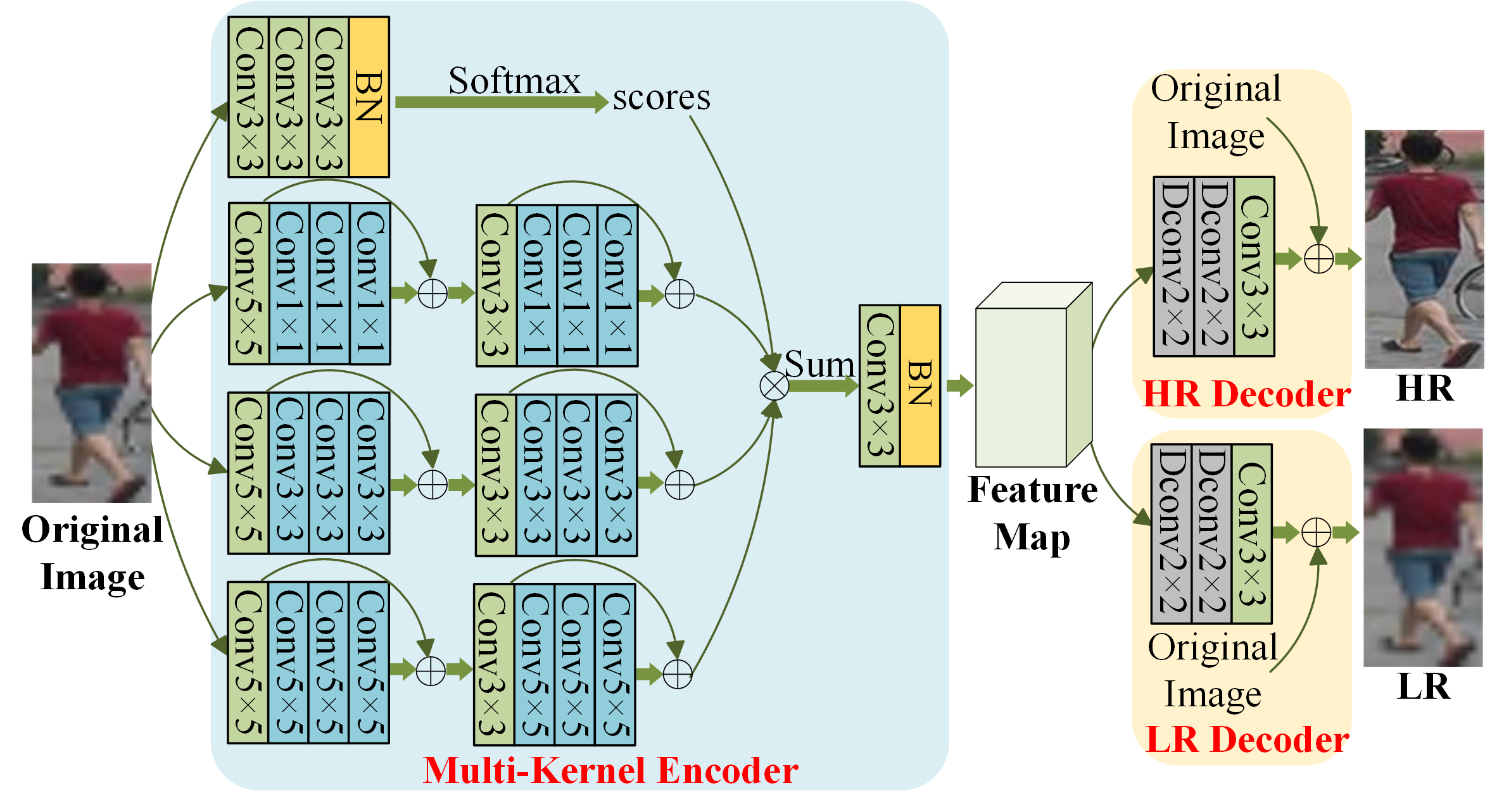}
\caption{The details of the Resolution Reconstruction Network.}
\label{fig3}
\end{figure}

\subsection{Cross-Resolution Person Re-ID}
To meet the challenge of cross-resolution person re-ID, a series of methods have been proposed and can be divided into two categories: 1) methods based on metric learning or dictionary learning and 2) methods based on SR. In the first category, Jing \emph{et al.} \cite{8} develop a semi-coupled low-rank dictionary learning approach to learn the mapping between HR and LR images. Li \emph{et al.} \cite{31} introduces a learning framework which jointly performs cross-scale image domain alignment and distance metric learning. However, the matching capability of these methods is limited due to the lack of fine-grained details in LR images.

The success of super-resolution (SR) technology promotes the development of cross-resolution person re-ID. The key idea of these methods is to restore LR images back to HR images by resolution reconstruction loss or GAN. Both \cite{10} and \cite{11} design a jointly learning framework which simultaneously optimize a SR model and a re-ID model. Wang \emph{et al.} \cite{12} present a cascaded structure to enhance image resolution step by step with the repeated use of SR-GAN \cite{34}. Li \emph{et al.} propose successively RAIN \cite{30} and CAD-Net \cite{29}. The former adopts GAN to generate resolution-invariant representations, while the latter adds the features extracted from recovered images and achieves better performance. Cheng \emph{et al.} \cite{32} introduce a training regularization method which utilizes the underlying association knowledge between SR and re-ID as an extra learning constraint to enhance the compatibility between two networks. Han \emph{et al.} \cite{43} propose an end-to-end PRI framework to adaptively predict the preferable scale factor, recover details for LR images and perform the identification. However, all above methods only focus on the HR features and neglect the useful information in LR images. In this work, we explore the influence of resolutions on feature extraction and verify that LR information matters for cross-resolution person re-ID. Based on the above idea, we develop a novel method fully utilizing features of different resolutions.

\section{Proposed Method}
\subsection{Framework Overview}
As illustrated in Figure \ref{fig2}, our proposed MRJL contains two sub-networks, RRN and DFFN. In the training phase, we define a set of input HR images with associated labels as ${D_H}{\rm{ = }}\left\{ {{x^H},y} \right\}$, where ${x^H} \in {\mathbb{R}^{H \times W \times 3}}$ represents a HR image and $y \in \mathbb{R}$ represents its identity label. To train the RRN with the capability to reconstruct different resolutions of images, we down-sample each HR image with the down-sampling rate $r \in \left\{ {2,3,4} \right\}$ (i.e., the spatial size of the down-sampled image becomes $\frac{H}{r} \times \frac{W}{r}$) and resize them back to the original size. The set of generated LR images obviously share the same identity labels and are denoted as ${D_L}{\rm{ = }}\left\{ {\left( {x_2^L,x_3^L,x_4^L} \right),y} \right\}$ where $x_i^L \in {\mathbb{R}^{H \times W \times 3}}$ is a LR image and the subscript $i \in \left\{ {2,3,4} \right\}$ represents the down-sampling rate. (The subscript $i$ is omitted in following paper for simplicity unless necessary.) In the testing phase, our framework regards the resolution of inputs as unknown, and processes the gallery (HR) and query (LR) equally.

In order to generate both HR and LR images for an input image with unknown resolution, we design the RRN module which is made up of an encoder and two independent decoders. The encoder is utilized to extract feature map from an input image, and the two decoders reconstruct the feature map into HR version and LR version, respectively. The DFFN module adopts a dual-branch structure to extract the feature representations ${f^H} \in {\mathbb{R}^d}$ and ${f^L} \in {\mathbb{R}^d}$ ($d$ denotes the dimension of feature) from the generated HR and LR images, respectively. Note that the two branches don’t share parameter weights. As for testing, feature representations $f^H$ and $f^L$ of all images in gallery and query sets are computed, and then the concatenation $f = \left[ {{f^H},{f^L}} \right] \in {\mathbb{R}^{2d}}$ will be used for distance measure.

\subsection{Resolution Reconstruction Network (RRN)}
Before feature extraction, the quality of generated images greatly affects the discrimination of representations. The proposed RRN module consists of a multi-kernel encoder, a HR decoder and a LR decoder, as shown in Figure \ref{fig3}.

The multi-kernel encoder ($ME$) has a four-branch structure that consists of three feature perception branches and an attention branch. All the perception branches are made up of 8 convolutional layers, and the attention branch has 3 convolutional layers followed by a batch normalization layer and a softmax activation function. To make the network perceive features of different scales, the kernel sizes of these perception branches are different, which are set to $\left\{ {1,3,5} \right\}$, respectively. Motivated by the previous works in SR \cite{11,35}, several skip connections are introduced to RRN to preserve the original visual cues and help reconstruct HR images.
Besides, attention mechanism \cite{36} has widely applied in neural network to make the network focus on parts of interest. In RRN, the attention branch is used to train the encoder to focus on the interested perceptual scale and then learn three attention weights for corresponding perception branches. The output feature map of the encoder is a weighted sum of all the outputs from the individual branch.

The HR decoder ($HD$) and LR decoder ($LD$) adopt the same network structure but don’t share parameter weights. Both decoders have 2 deconvolution layers and 1 convolution layers. For each input image $x$ (It doesn’t matter whether it is ${x^H}$ or ${x^L}$), our RRN can reconstruct both HR and LR images as:
\begin{equation}
    {\tilde x^H} = HD\left( {ME\left( x \right)} \right),{\tilde x^L} = LD\left( {ME\left( x \right)} \right)
\end{equation}

According to the formula above, if the training input is a HR image ${x^H}$, its reconstructed HR version and reconstructed LR version are denoted as ${\tilde x^{H2H}}$ and ${\tilde x^{H2L}}$, respectively. Similarly, if the training input is a LR image ${x^L}$, its two reconstructed versions are denoted as $\tilde x_i^{L2H}$ and $\tilde x_i^{L2L}$, where the subscript $i \in \left\{ {2,3,4} \right\}$ represents the corresponding down-sampling rate.

\begin{figure}[t]
\centering
\includegraphics[width=0.9\columnwidth]{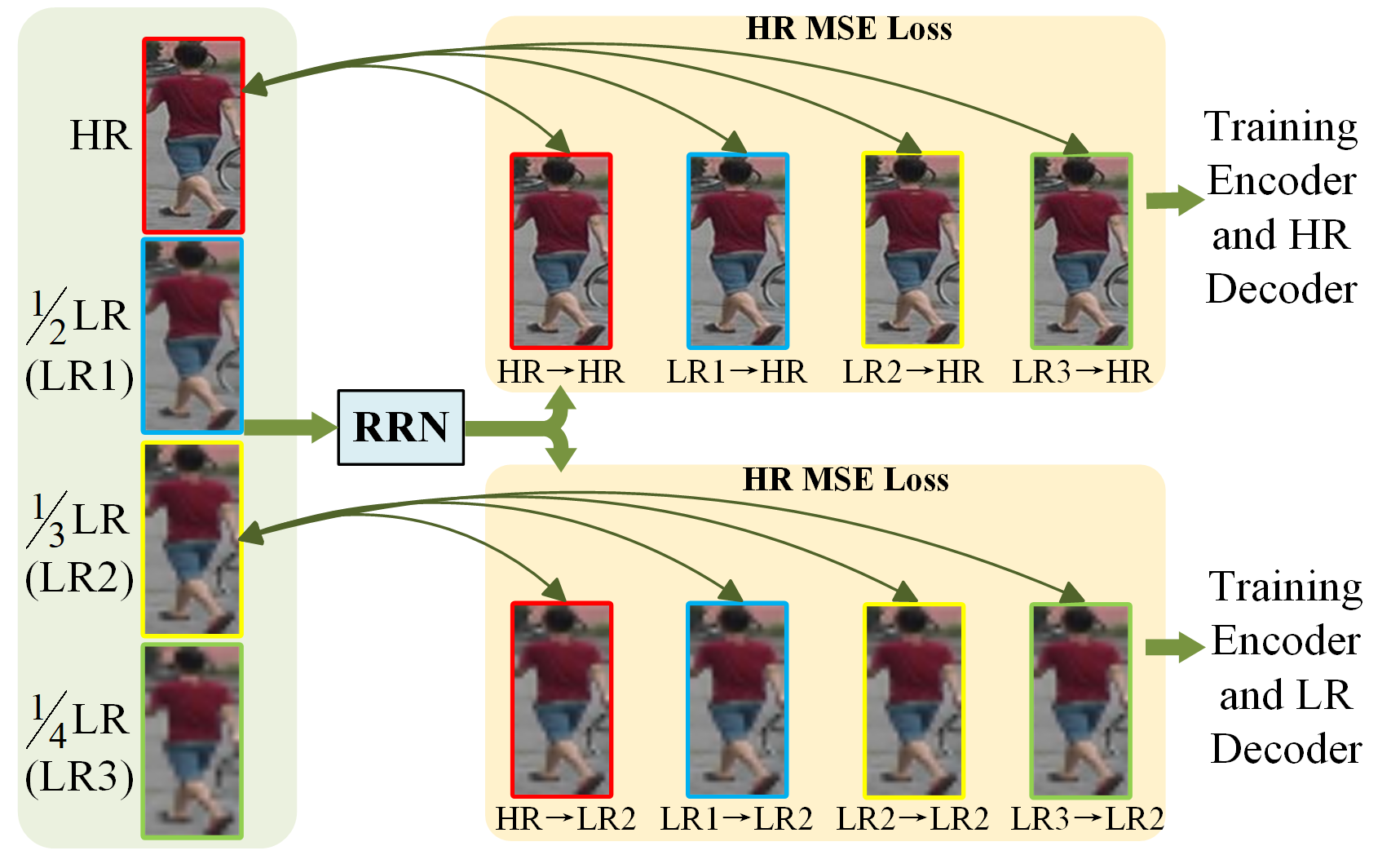}
\caption{The training strategy of RRN, consists of two aspects: 1) training the reconstructed HR images to be visually closer to the original HR images, and 2) training the reconstructed LR images to be visually closer to the standard LR images.}
\label{fig4}
\end{figure}

As illustrated in Figure \ref{fig4}, pixel-wise Mean Square Error (MSE) loss \cite{37} is applied in the training strategy of RRN which simultaneously trains the encoder and two different decoders. Since these LR images have variable resolutions with different down-sampling rates picked from $\left\{ {2,3,4} \right\}$, we should set a LR reference standard for RRN. Here we select the median  $x_3^L$ in LR images as the standard. To train the encoder and decoders, the HR MSE loss $L_{mse}^H$ and the LR MSE loss $L_{mse}^L$ are calculated as:
\begin{equation}
    L_{mse}^H = \left\| {{{\tilde x}^{H2H}} - {x^H}} \right\|_2^2 + \sum\limits_{i = 2}^4 {\left\| {\tilde x_i^{L2H} - {x^H}} \right\|_2^2}
\end{equation}
\begin{equation}
    L_{mse}^L = \left\| {{{\tilde x}^{H2L}} - x_3^L} \right\|_2^2 + \sum\limits_{i = 2}^4 {\left\| {\tilde x_i^{L2L} - x_3^L} \right\|_2^2}
\end{equation}
Then we get the joint MSE loss in RRN as:
\begin{equation}
    {L_{mse}} = L_{mse}^{LR} + \lambda L_{mse}^{HR}
\end{equation}
where $\lambda$ is a hyper-parameter to control the importance of HR MSE loss.

\subsection{Dual Feature Fusion Network (DFFN)}
A dual-branch structure is used in DFFN to extract both detailed information from HR images and complementary information in LR images simultaneously. Two branches in DFFN share the same network but do not share parameter weights since we wish each branch to focus on different types of features from images with different resolution.
Here we adopt the feature extractor on the basis of PCB method \cite{2}. The 3D tensor generated by the backbone network (e.g., ResNet50) is segmented into 4 horizontal stripes. Followed by an average pooling layer and $1 \times 1$ convolutional layers, the feature representation ${f^H} = \left[ {f_1^H,f_2^H,f_3^H,f_4^H,f_5^H} \right]$ (or ${f^L} = \left[ {f_1^L,f_2^L,f_3^L,f_4^L,f_5^L} \right]$) is obtained which is concatenated by 4 256-dimentional local features and a 512-dimentional global feature.

In DFFN, we adopt both triplet loss and cross entropy loss to enhance the discrimination of feature representations:
\begin{equation}
    {L_{reid}} = {L_{ce}} + \gamma {L_{trip}}
\end{equation}
where $\gamma$ is a hyper-parameter to control the importance of triplet loss. The cross entropy loss ${L_{ce}}$ can be computed as:
\begin{equation}
{L_{ce}} =  - \sum\limits_{i = 1}^5 {\left( {y\log \left( {FC\left( {f_i^H} \right)} \right) + y\log \left( {FC\left( {f_i^L} \right)} \right)} \right)}
\end{equation}
where $y$ denotes the ground truth and the predicted person label can be generated by FC layers. The triplet loss ${L_{trip}}$ can be calculated as:
\begin{equation}
    \begin{array}{l}
{L_{trip}} = \sum\limits_{i = 1}^5 {\sum\limits_{f_{a,i}^H,f_{p,i}^H,f_{n,i}^H} {{{\left[ {d\left( {f_{a,i}^H,f_{p,i}^H} \right) - d\left( {f_{a,i}^H,f_{n,i}^H} \right) + m} \right]}_ + }} } \\
{\kern 1pt} {\kern 1pt} {\kern 1pt} {\kern 1pt} {\kern 1pt} {\kern 1pt} {\kern 1pt} {\kern 1pt} {\kern 1pt} {\kern 1pt} {\kern 1pt} {\kern 1pt} {\kern 1pt} {\kern 1pt} {\kern 1pt} {\kern 1pt} {\kern 1pt} {\kern 1pt} {\kern 1pt} {\kern 1pt} {\kern 1pt} {\kern 1pt} {\kern 1pt} {\kern 1pt} {\kern 1pt} {\rm{ + }}\sum\limits_{i = 1}^5 {\sum\limits_{f_{a,i}^L,f_{p,i}^L,f_{n,i}^L} {{{\left[ {d\left( {f_{a,i}^L,f_{p,i}^L} \right) - d\left( {f_{a,i}^L,f_{n,i}^L} \right) + m} \right]}_ + }} } 
\end{array}
\end{equation}
where $f_{a,i}^H$, $f_{p,i}^H$ and $f_{n,i}^H$ indicate the ${i^{{\rm{th}}}}$ sub-features extracted from anchor, positive and negative HR samples ($f_{a,i}^L$, $f_{p,i}^L$ and $f_{n,i}^L$ indicate the corresponding features from LR samples), $d\left( { \cdot , \cdot } \right)$ indicates the Euclidean distance, and $m$ is a margin hyper-parameter to control the differences between intra and inter distances.

\section{Experiment}

\begin{table*}[t]
\centering
\resizebox{\textwidth}{!}{
\begin{tabular}{l c c c c c c c c c c c}
    \hline
    \multirow{2}{*}{Method} & \multirow{2}{*}{Publication} & \multicolumn{2}{c}{MLR-SYSU} & \multicolumn{2}{c}{MLR-VIPER} & \multicolumn{2}{c}{MLR-CUHK03} & \multicolumn{2}{c}{MLR-Market-1501} & \multicolumn{2}{c}{CAVIAR} \\
    \cline{3-4}
    \cline{5-6}
    \cline{7-8}
    \cline{9-10}
    \cline{11-12}
       & & Rank-1 & Rank-5 & Rank-1 & Rank-5 & Rank-1 & Rank-5 & Rank-1 & Rank-5 & Rank-1 & Rank-5 \\
    \hline
    JUDEA \cite{31} & ICCV'15 & 18.3 & 41.9 & 26.0 & 55.1 & 26.2 & 58.0 & - & - & 22.0 & 60.1 \\
    SLD$^{2}$L \cite{8} & CVPR'15 & 20.3 & 34.8 & 20.3 & 44.0 & - & - & - & - & 18.4 & 44.8 \\
    SDF \cite{9} & IJCAI'16 & 13.3 & 26.7 & 9.3 & 38.1 & 22.2 & 48.0 & - & - & 14.3 & 37.5 \\
    SING \cite{10} & AAAI'18 & \textcolor{blue}{\textbf{50.7}} & \textcolor{blue}{\textbf{75.4}} & 33.5 & 57.0 & 67.7 & 90.7 & 74.4 & 87.8 & 33.5 & 72.7 \\
    CSR-GAN \cite{12} & IJCAI'18 & - & - & 37.2 & 62.3 & 70.7 & 92.1 & 76.4 & 88.5 & 32.3 & 70.9 \\
    FFSR+RIFE \cite{11} & IJCAI'19 & - & - & 41.6 & 64.9 & 73.3 & 92.6 & - & - & 36.4 & 72.0 \\
    RAIN \cite{30} & AAAI'19 & - & - & 42.5 & 68.3 & 78.9 & 97.3 & - & - & 42.0 & 77.3 \\
    CDA-Net \cite{29} & ICCV'19 & - & - & 43.1 & 68.2 & 82.1 & 97.4 & 83.7 & 92.7 & 42.8 & 76.2 \\
    PCB+PRI \cite{43} & ECCV'20 & - & - & - & - & 86.2 & \textcolor{red}{\textbf{97.9}} & 88.1 & 94.2 & \textcolor{blue}{\textbf{44.3}} & \textcolor{red}{\textbf{83.7}} \\
    INTACT \cite{32} & CVPR'20 & - & - & \textcolor{blue}{\textbf{46.2}} & \textcolor{blue}{\textbf{73.1}} & \textcolor{blue}{\textbf{86.4}} & \textcolor{blue}{\textbf{97.4}} & \textcolor{blue}{\textbf{88.1}} & \textcolor{blue}{\textbf{95.0}} & 44.0 & 81.8 \\
    \hline
    MRJL (Ours) & & \textcolor{red}{\textbf{73.0}} & \textcolor{red}{\textbf{87.3}} & \textcolor{red}{\textbf{58.7}} & \textcolor{red}{\textbf{84.1}} & \textcolor{red}{\textbf{90.7}} & 95.7 & \textcolor{red}{\textbf{90.1}} & \textcolor{red}{\textbf{95.6}} & \textcolor{red}{\textbf{61.2}} & \textcolor{blue}{\textbf{82.4}} \\
    \hline
\end{tabular}}
\caption{Comparisons of our proposed method to the state-of-the-arts (\%). \textcolor{red}{Red} and \textcolor{blue}{blue} bold numbers indicate the ${1^{{\rm{st}}}}$ and ${2^{{\rm{nd}}}}$ top results.}
\label{tab1}
\end{table*}

\subsection{Datasets}
Five person re-ID datasets are used to evaluate our proposed method, including four synthetic Multiple Low Resolutions (MLR) datasets and one real-world dataset. The generation strategy of MLR datasets refers \cite{10,29,32}. Specifically, we down-sample images from one camera by randomly selecting a down-sampling rate $r \in \left\{ {2,3,4} \right\}$, while the images captured by other camera(s) remain unchanged.
1) \textbf{MLR-SYSU} is constructed from the SYSU \cite{38}. SYSU contains 502 identities captured by 2 cameras, and three images per person are randomly selected for each camera. Half of these identities are for training and half are for testing.
2)	\textbf{MLR-VIPeR} is a synthetic version built from the VIPeR \cite{39}. VIPeR contains 632 person image pairs taken by 2 cameras. According to the identity labels, these pairs are divided into 2 non-overlapping halves.
3)	\textbf{MLR-CUHK03} is based on the CUHK03 \cite{40}. CUHK03 is composed of five different pairs of camera views, and has 14,097 images of 1,467 identities. We use the 1367/100 training/testing identity split. 
4)	\textbf{MLR-Market-1501} is built from the Market-1501 \cite{41}. Market-1501 comprises more than 32,000 images of 1,501 identities from 6 cameras, and we utilize 751/750 training/testing identity split.
5)	\textbf{CAVIAR} \cite{42} is a challenging real-world person re-ID dataset which contains 1220 images of 72 identities captured by 2 cameras. Among them, 22 persons who appear only in the close camera are discarded. Similar to MLR-VIPeR, we split the remaining images into 2 non-overlapping halves.

\subsection{Implementation Details}
In the training phase, all the input images are resized to $128 \times 256$. A mini-batch has 20 images of 5 persons where 2 HR images (each HR image can generate 3 down-sampled LR versions) and 2 original LR images are selected for each person. Noting that original LR samples are only utilized to train the DFFN module. Hyper-parameters $\lambda$, $\gamma$ and $m$ are set to 100, 1, 0.5, respectively. We select Adam to optimize our model with weight decay $5 \times {10^{ - 4}}$. For parameters in the MSE loss, we set a learning rate of $3 \times {10^{ - 3}}$, and for parameters in the re-ID loss, we set a learning rate of $3 \times {10^{ - 4}}$. Our model is trained for 60 epochs in total, and the learning rates are decreased by 0.1 after 30 epochs. 

\begin{figure}[t]
\centering
\includegraphics[width=0.9\columnwidth]{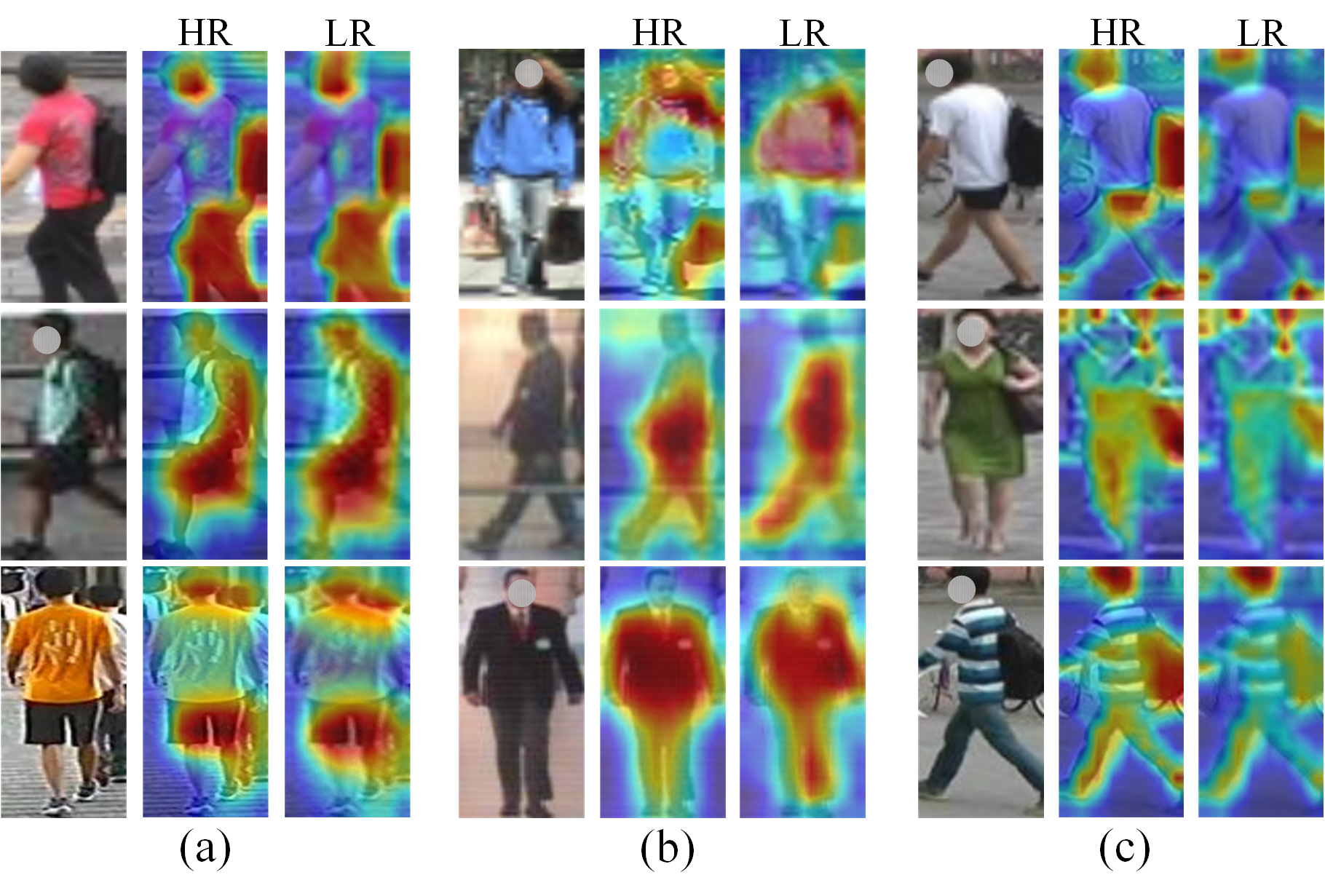}
\caption{Examples of feature response maps extracted on different resolution samples. All the cases are classified into three groups.} 
\label{fig5}
\end{figure}

\subsection{Comparison with State-of-the-art Approaches}

We compare our method with several recent cross-resolution person re-ID methods, and the comparable results are reported in Table \ref{tab1}, which show that the SR based methods commonly achieve better performance than the traditional methods. One important reason is that these SR based methods aid in the recovery of missing spatial information that contains more discriminative features. In contrast, traditional methods are incapable of recovering the lost information, resulting in poor performance.
From Table \ref{tab1}, we can also observe that our proposed MRJL outperforms the state-of-the-arts by \textbf{22.3\%}, \textbf{12.5\%}, \textbf{4.3\%}, \textbf{2.0\%} and \textbf{17.2\%} in Rank-1 on MLR-SYSU, MLR-VIPeR, MLR-CUHK03, MLR-Market-1501 and CAVIAR, respectively. The performance superiority of our method can be mainly attributed to the joint representations of both HR and LR features. All existing SR based methods only extract features from recovered HR images but ignore the complementary information provided by LR ones.


\subsection{Ablation Study}

\subsubsection{Influence of Resolutions on Feature Extraction}
To investigate the influence of resolution on feature extraction, we conduct the following experiments as shown in Table \ref{tab2}. The variant (1.3) extracts the joint HR and LR feature representations, while the variant (1.1) and variant (1.2) only utilize the single branch in RRN and DFFN. The comparison results confirm two assumptions: 1) LR information also matters for cross-resolution person re-ID. LR features can provide complementary information for HR features and further improve the accuracy of matching. 2) Compared with HR images, networks can extract more discriminative features from LR images in some cases, such as on MLR-SYSU and CAVIAR datasets.

\begin{table}[t]
\centering
\resizebox{\columnwidth}{!}{
\begin{tabular}{l c c c c c}
    \hline
    \multirow{2}{*}{Resolution} & \multicolumn{4}{c}{MLR-Datasets (Rank-1)} & CAVIAR \\
    \cline{2-5}
     & SYSU & VIPeR & CUHK03 & Market-1501 & (Rank-1) \\
    \hline
    (1.1) HR & 68.0 & 54.0 & \textbf{90.7} & 88.9 & 50.1 \\
    (1.2) LR & 70.0 & 48.9 & 88.8 & 88.4 & 53.6 \\
    (1.3) HR+LR & \textbf{73.0} & \textbf{58.7} & \textbf{90.7} & \textbf{90.1} & \textbf{61.2} \\
    \hline
\end{tabular}}
\caption{Effects of different resolutions(\%).}
\label{tab2}
\end{table}

Figure \ref{fig5} visualizes some feature response maps extracted from different resolution samples which further verify the above viewpoints. We classify the different cases into 3 groups. Most cases are similar to the group (a), which reflects that the network can extract similar features from HR and LR images. Group (b) shows that the network extract more global information from LR images compared with HR ones, and group (c) indicates that HR images make it easier for the network to focus on detail information, such as bags and textures. These experiments can provide a reasonable explanation for the pending phenomenon mentioned in \cite{11} that re-ID model achieves lower accuracy when the recovered images become higher resolution. Although the recovered images obtain better visual quality, they have higher risk to generate false details which may mislead feature extraction. In most cases, LR images can still provide discriminative features for matching.

\begin{figure}[t]
\centering
\includegraphics[width=0.9\columnwidth]{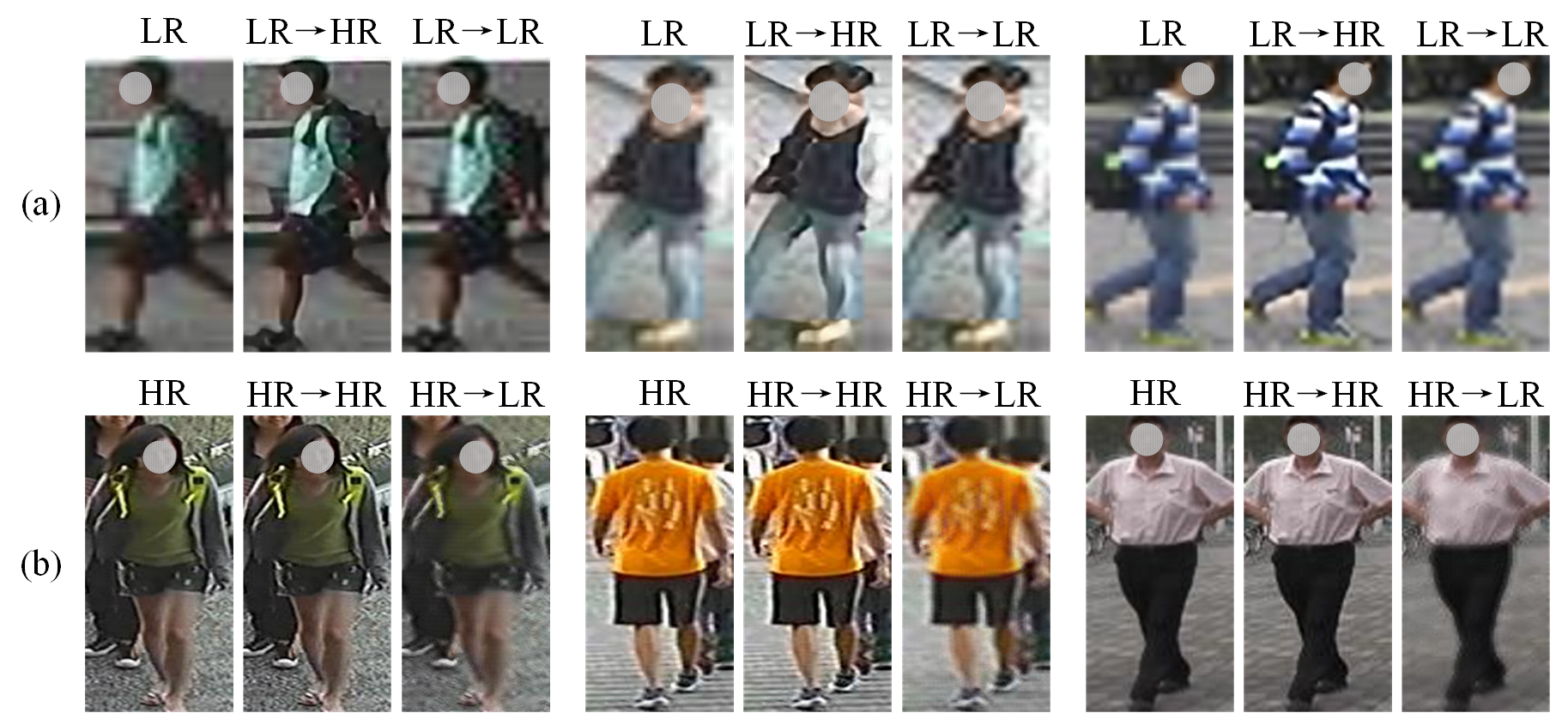}
\caption{Visual results of the reconstructed HR and LR images. The group (a) and (b) represent the situations that the input is a LR (query) or HR (gallery) image, respectively.}
\label{fig6}
\end{figure}

\begin{table}[t]
\centering
\resizebox{\columnwidth}{!}{
\begin{tabular}{l c c c c c}
    \hline
    \multirow{2}{*}{Encoder Structure} & \multicolumn{4}{c}{MLR-Datasets (Rank-1)} & CAVIAR \\
    \cline{2-5}
     & SYSU & VIPeR & CUHK03 & Market-1501 & (Rank-1) \\
    \hline
    (2.1) Single-Branch & 71.9 & 57.8 & 90.6 & 89.8 & 54.0 \\
    (2.2) Multi-Kernel & \textbf{73.0} & \textbf{58.7} & \textbf{90.7} & \textbf{90.1} & \textbf{61.2} \\
    \hline
\end{tabular}}
\caption{Effects of different encoder structures in RRN (\%).}
\label{tab3}
\end{table}

\begin{figure}[t]
\centering
\includegraphics[width=0.9\columnwidth]{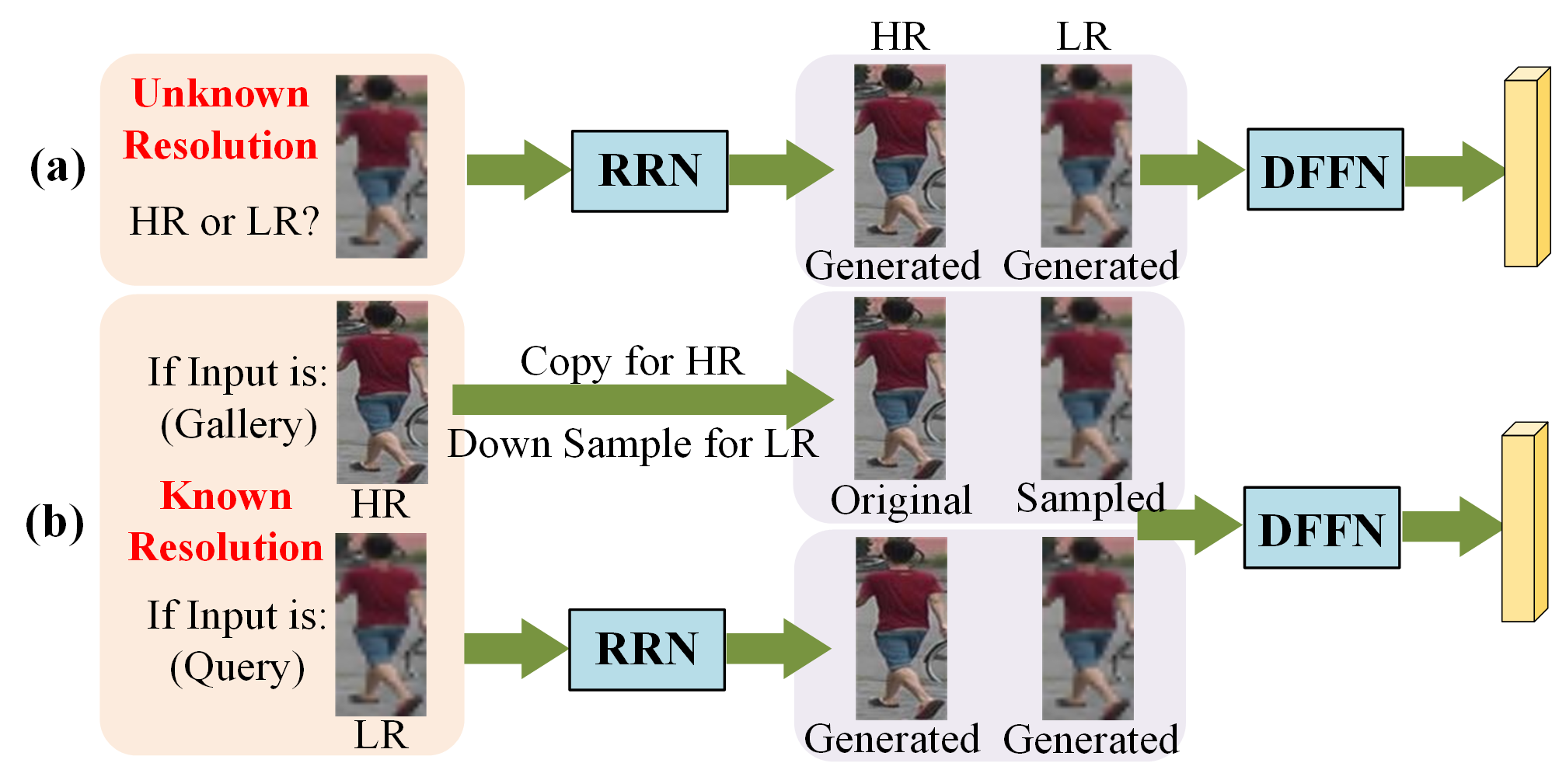}
\caption{The method structures of (un)known resolution situations.}
\label{fig7}
\end{figure}

\subsubsection{Analysis on RRN}
To evaluate the validity of RRN module, we conduct the following ablation experiments. Firstly, we compare the single-branch encoder (variant (2.1)) and our multi-kernel encoder (variant (2.2)), and the comparable results are listed in Table \ref{tab3}. It can be observed that, compared with variant (2.1), variant (2.2) achieves higher accuracy on all datasets. The results prove the effectiveness of multi-kernel structure which can perceive features of different scales.

In addition, we test the different LR standards in the training strategy of RRN. During the training phase of RRN, HR images have the definite standard that the original images without down-sampling. Nevertheless, the LR images have variable down-sampling rates which mean different reference standards for reconstructing LR images, as shown in Figure \ref{fig4}. Table \ref{tab4} reports that it is best to choose LR images with down-sampling rate 3 as the LR reference standard in most datasets except for MLR-CUHK03. The LR standard determines the resolution of reconstructed LR images. The results indicate that the discrimination of LR features will decrease if the reconstructed LR images are too close to HR images or too vague to mine features. Note that MLR-CUHK03 achieves the best performance when the LR standard is $x_4^L$. It means $x_3^L$ cannot provide enough complementary information on MLR-CUHK03, which is also reflected on Table \ref{tab2}, and the better LR feature extraction needs lower resolution of reconstructed LR images. One possible explanation is that the samples in MLR-CUHK03 is relatively clearer, which limit the advantages of LR complementary effects.

To further verify the reconstruction capability of RRN, we visualize the reconstructed images. The two groups in Figure \ref{fig6} reflect that RRN is capable of generating HR and LR images regardless the resolution of the input image.

\begin{table}[t]
\centering
\resizebox{\columnwidth}{!}{
\begin{tabular}{l c c c c c}
    \hline
    \multirow{2}{*}{Standard} & \multicolumn{4}{c}{MLR-Datasets (Rank-1)} & CAVIAR \\
    \cline{2-5}
     & SYSU & VIPeR & CUHK03 & Market-1501 & (Rank-1) \\
    \hline
    (3.1) $x_2^L$ & 72.8 & 58.1 & 90.4 & 90.0 & 57.2 \\
    (3.2) $x_3^L$ & \textbf{73.0} & \textbf{58.7} & 90.7 & \textbf{90.1} & \textbf{61.2} \\
    (3.3) $x_4^L$ & 71.7 & 55.9 & \textbf{92.3} & 89.9 & 57.6 \\
    \hline
\end{tabular}}
\caption{Effects of different LR reference standards in RRN (\%).}
\label{tab4}
\end{table}

\subsubsection{Analysis on Unknown Resolution Strategy}

We divide the cross-resolution person re-ID into two categories: unknown resolution and known resolution, as illustrated in Figure \ref{fig7}. In the case of known situation, we assume the gallery images are all HR and the query images are all LR, and SR networks only need to pre-process the LR images. In the other case, we treat all the images equally in both gallery and query sets since networks needn’t know image
resolutions. Compared with known resolution case, unknown resolution case has two advantages: 1) Unknown resolution methods don’t need the resolution labels which are hard to quantify with pixel size. 2) The unknown resolution case is closer to the unconstrained scenarios, because resolutions of images will not be divided neatly in some practical applications. Table \ref{tab5} reports that our unknown resolution structure (variant (4.2)) achieves a few improvements, which can be explained that a few LR gallery images are mistaken for HR due to the classification by pixel size in structure (b).

\begin{table}[H]
\centering
\resizebox{\columnwidth}{!}{
\begin{tabular}{l c c c c c}
    \hline
    \multirow{2}{*}{Method} & \multicolumn{4}{c}{MLR-Datasets (Rank-1)} & CAVIAR \\
    \cline{2-5}
     & SYSU & VIPeR & CUHK03 & Market-1501 & (Rank-1) \\
    \hline
    (4.1) Known & 72.6 & 56.5 & 90.6 & 89.6 & 60.8 \\
    (4.2) Unknown & \textbf{73.0} & \textbf{58.7} & \textbf{90.7} & \textbf{90.1} & \textbf{61.2} \\
    \hline
\end{tabular}}
\caption{Effects of different method structures(\%).}
\label{tab5}
\end{table}

\section{Conclusion}
In this paper, we have investigated into the influence of resolutions on feature extraction, and proposed a Multi-Resolution Representation Joint Learning (MRJL) method to solve the cross-resolution person re-ID problem. By a series of experiments, we explore the effectiveness of LR features which is capable of complementing HR features. According to the inspiration, the MRJL utilizes a Resolution Reconstruction Network (RRN) to generate both HR and LR versions no matter what the input resolution is. Besides, a Dual Feature Fusion Network (DFFN) is designed to extract discriminative multi-resolution representations. Extensive experimental results on five challenging datasets demonstrate the superiority of the MRJL over the relevant state-of-the-art methods.

\section*{Acknowledgements}
This research was conducted in collaboration with Singapore Telecommunications Limited and supported by the Singapore Government through the Industry Alignment Fund - Industry Collaboration Projects Grant (No. NTU 2018-0551).

\bibliographystyle{named}
\bibliography{ijcai21}

\begin{thebibliography}{}

\bibitem[\protect\citeauthoryear{Chen \bgroup \em et al.\egroup }{2016}]{38}
Ying-Cong Chen, Wei-Shi Zheng, Jian-Huang Lai, and Pong~C Yuen.
\newblock An asymmetric distance model for cross-view feature mapping in person
  reidentification.
\newblock {\em IEEE Trans. Circuits Syst. Video Technol}, 27(8):1661--1675,
  2016.

\bibitem[\protect\citeauthoryear{Chen \bgroup \em et al.\egroup }{2019}]{30}
Yun-Chun Chen, Yu-Jhe Li, Xiaofei Du, and Yu-Chiang~Frank Wang.
\newblock Learning resolution-invariant deep representations for person
  re-identification.
\newblock In {\em AAAI}, volume~33, pages 8215--8222, 2019.

\bibitem[\protect\citeauthoryear{Cheng \bgroup \em et al.\egroup }{2011}]{42}
Dong~Seon Cheng, Marco Cristani, Michele Stoppa, Loris Bazzani, and Vittorio
  Murino.
\newblock Custom pictorial structures for re-identification.
\newblock In {\em BMVC}, volume~1, page~6, 2011.

\bibitem[\protect\citeauthoryear{Cheng \bgroup \em et al.\egroup }{2020}]{32}
Zhiyi Cheng, Qi~Dong, Shaogang Gong, and Xiatian Zhu.
\newblock Inter-task association critic for cross-resolution person
  re-identification.
\newblock In {\em CVPR}, pages 2605--2615, 2020.

\bibitem[\protect\citeauthoryear{Dong \bgroup \em et al.\egroup }{2015}]{37}
Chao Dong, Chen~Change Loy, Kaiming He, and Xiaoou Tang.
\newblock Image super-resolution using deep convolutional networks.
\newblock {\em IEEE Trans. Pattern Anal. Mach. Intell}, 38(2):295--307, 2015.

\bibitem[\protect\citeauthoryear{Gray and Tao}{2008}]{39}
Douglas Gray and Hai Tao.
\newblock Viewpoint invariant pedestrian recognition with an ensemble of
  localized features.
\newblock In {\em ECCV}, pages 262--275. Springer, 2008.

\bibitem[\protect\citeauthoryear{Han \bgroup \em et al.\egroup }{2020}]{43}
Ke~Han, Yan Huang, Zerui Chen, Liang Wang, and Tieniu Tan.
\newblock Prediction and recovery for adaptive low-resolution person
  re-identification.
\newblock In {\em ECCV}, pages 193--209. Springer, 2020.

\bibitem[\protect\citeauthoryear{Jiao \bgroup \em et al.\egroup }{2018}]{10}
Jiening Jiao, Wei-Shi Zheng, Ancong Wu, Xiatian Zhu, and Shaogang Gong.
\newblock Deep low-resolution person re-identification.
\newblock In {\em AAAI}, 2018.

\bibitem[\protect\citeauthoryear{Jing \bgroup \em et al.\egroup }{2015}]{8}
Xiao-Yuan Jing, Xiaoke Zhu, Fei Wu, Xinge You, Qinglong Liu, Dong Yue, Ruimin
  Hu, and Baowen Xu.
\newblock Super-resolution person re-identification with semi-coupled low-rank
  discriminant dictionary learning.
\newblock In {\em CVPR}, pages 695--704, 2015.

\bibitem[\protect\citeauthoryear{Kalayeh \bgroup \em et al.\egroup }{2018}]{4}
Mahdi~M Kalayeh, Emrah Basaran, Muhittin G{\"o}kmen, Mustafa~E Kamasak, and
  Mubarak Shah.
\newblock Human semantic parsing for person re-identification.
\newblock In {\em CVPR}, pages 1062--1071, 2018.

\bibitem[\protect\citeauthoryear{Ledig \bgroup \em et al.\egroup }{2017}]{34}
Christian Ledig, Lucas Theis, Ferenc Husz{\'a}r, Jose Caballero, Andrew
  Cunningham, Alejandro Acosta, Andrew Aitken, Alykhan Tejani, Johannes Totz,
  Zehan Wang, et~al.
\newblock Photo-realistic single image super-resolution using a generative
  adversarial network.
\newblock In {\em CVPR}, pages 4681--4690, 2017.

\bibitem[\protect\citeauthoryear{Li \bgroup \em et al.\egroup }{2014}]{40}
Wei Li, Rui Zhao, Tong Xiao, and Xiaogang Wang.
\newblock Deepreid: Deep filter pairing neural network for person
  re-identification.
\newblock In {\em CVPR}, pages 152--159, 2014.

\bibitem[\protect\citeauthoryear{Li \bgroup \em et al.\egroup }{2015}]{31}
Xiang Li, Wei-Shi Zheng, Xiaojuan Wang, Tao Xiang, and Shaogang Gong.
\newblock Multi-scale learning for low-resolution person re-identification.
\newblock In {\em ICCV}, pages 3765--3773, 2015.

\bibitem[\protect\citeauthoryear{Li \bgroup \em et al.\egroup }{2018}]{22}
Wei Li, Xiatian Zhu, and Shaogang Gong.
\newblock Harmonious attention network for person re-identification.
\newblock In {\em CVPR}, pages 2285--2294, 2018.

\bibitem[\protect\citeauthoryear{Li \bgroup \em et al.\egroup }{2019}]{29}
Yu-Jhe Li, Yun-Chun Chen, Yen-Yu Lin, Xiaofei Du, and Yu-Chiang~Frank Wang.
\newblock Recover and identify: A generative dual model for cross-resolution
  person re-identification.
\newblock In {\em ICCV}, pages 8090--8099, 2019.

\bibitem[\protect\citeauthoryear{Liu \bgroup \em et al.\egroup }{2018}]{23}
Jinxian Liu, Bingbing Ni, Yichao Yan, Peng Zhou, Shuo Cheng, and Jianguo Hu.
\newblock Pose transferrable person re-identification.
\newblock In {\em CVPR}, pages 4099--4108, 2018.

\bibitem[\protect\citeauthoryear{Mao \bgroup \em et al.\egroup }{2016}]{35}
Xiaojiao Mao, Chunhua Shen, and Yu-Bin Yang.
\newblock Image restoration using very deep convolutional encoder-decoder
  networks with symmetric skip connections.
\newblock In {\em NIPS}, pages 2802--2810, 2016.

\bibitem[\protect\citeauthoryear{Mao \bgroup \em et al.\egroup }{2019}]{11}
Shunan Mao, Shiliang Zhang, and Ming Yang.
\newblock Resolution-invariant person re-identification.
\newblock {\em arXiv preprint arXiv:1906.09748}, 2019.

\bibitem[\protect\citeauthoryear{Sun \bgroup \em et al.\egroup }{2018}]{2}
Yifan Sun, Liang Zheng, Yi~Yang, Qi~Tian, and Shengjin Wang.
\newblock Beyond part models: Person retrieval with refined part pooling (and a
  strong convolutional baseline).
\newblock In {\em ECCV}, pages 480--496, 2018.

\bibitem[\protect\citeauthoryear{Vaswani \bgroup \em et al.\egroup }{2017}]{36}
Ashish Vaswani, Noam Shazeer, Niki Parmar, Jakob Uszkoreit, Llion Jones,
  Aidan~N Gomez, {\L}ukasz Kaiser, and Illia Polosukhin.
\newblock Attention is all you need.
\newblock {\em NIPS}, 30:5998--6008, 2017.

\bibitem[\protect\citeauthoryear{Wang \bgroup \em et al.\egroup }{2016}]{9}
Zheng Wang, Ruimin Hu, Yi~Yu, Junjun Jiang, Chao Liang, and Jinqiao Wang.
\newblock Scale-adaptive low-resolution person re-identification via learning a
  discriminating surface.
\newblock In {\em IJCAI}, volume~2, page~6, 2016.

\bibitem[\protect\citeauthoryear{Wang \bgroup \em et al.\egroup }{2018a}]{3}
Guanshuo Wang, Yufeng Yuan, Xiong Chen, Jiwei Li, and Xi~Zhou.
\newblock Learning discriminative features with multiple granularities for
  person re-identification.
\newblock In {\em ACM Multimedia}, pages 274--282, 2018.

\bibitem[\protect\citeauthoryear{Wang \bgroup \em et al.\egroup }{2018b}]{12}
Zheng Wang, Mang Ye, Fan Yang, Xiang Bai, and Shin'ichi Satoh.
\newblock Cascaded sr-gan for scale-adaptive low resolution person
  re-identification.
\newblock In {\em IJCAI}, volume~1, page~4, 2018.

\bibitem[\protect\citeauthoryear{Yoo \bgroup \em et al.\egroup }{2015}]{44}
Donggeun Yoo, Sunggyun Park, Joon-Young Lee, and In~So~Kweon.
\newblock Multi-scale pyramid pooling for deep convolutional representation.
\newblock In {\em CVPR}, pages 71--80, 2015.

\bibitem[\protect\citeauthoryear{Zhao \bgroup \em et al.\egroup }{2017}]{15}
Haiyu Zhao, Maoqing Tian, Shuyang Sun, Jing Shao, Junjie Yan, Shuai Yi,
  Xiaogang Wang, and Xiaoou Tang.
\newblock Spindle net: Person re-identification with human body region guided
  feature decomposition and fusion.
\newblock In {\em CVPR}, pages 1077--1085, 2017.

\bibitem[\protect\citeauthoryear{Zheng \bgroup \em et al.\egroup }{2015}]{41}
Liang Zheng, Liyue Shen, Lu~Tian, Shengjin Wang, Jingdong Wang, and Qi~Tian.
\newblock Scalable person re-identification: A benchmark.
\newblock In {\em ICCV}, pages 1116--1124, 2015.

\bibitem[\protect\citeauthoryear{Zheng \bgroup \em et al.\egroup }{2016}]{1}
Liang Zheng, Yi~Yang, and Alexander~G Hauptmann.
\newblock Person re-identification: Past, present and future.
\newblock {\em arXiv preprint arXiv:1610.02984}, 2016.

\end{thebibliography}

\end{document}